\documentclass[lettersize,journal]{IEEEtran}
\usepackage{amsmath,amsfonts}
\usepackage{algorithmic}
\usepackage{algorithm}
\usepackage{array}
\usepackage[caption=false,font=normalsize,labelfont=sf,textfont=sf]{subfig}
\usepackage{textcomp}
\usepackage{stfloats}
\usepackage{url}
\usepackage{verbatim}
\usepackage{graphicx}
\usepackage{cite}
\usepackage{amsthm}
\usepackage{booktabs}
\usepackage[switch]{lineno}
\usepackage{multirow}
\usepackage{diagbox}
\usepackage{svg}
\usepackage{amssymb}
\usepackage{tabularx}
\usepackage{siunitx}
\usepackage{rotating}
\usepackage{enumitem}
\usepackage{newfloat}
\usepackage{listings}
\usepackage{adjustbox}

\definecolor{lightgreen}{RGB}{220,240,220}

\newcolumntype{C}[1]{>{\centering\arraybackslash}m{#1}}

\newcolumntype{R}[1]{>{\raggedleft\arraybackslash}p{#1}} % Define a new column type

\begin{document}

\title{MF-CLIP: Leveraging CLIP as Surrogate Models for No-box Adversarial Attacks}

\author{Jiaming Zhang, Lingyu Qiu, Qi Yi, Yige Li, Jitao Sang,\\
Changsheng Xu,~\IEEEmembership{Fellow, ~IEEE}, and Dit-Yan Yeung,~\IEEEmembership{Senior Member, ~IEEE}
        % <-this % stops a space

\thanks{Jiaming Zhang and Dit-Yan Yeung are with the Department of Computer Science and Engineering, Hong Kong University of Science and Technology, Hong Kong, China (e-mail:jmzhang@ust.hk, dyyeung@cse.ust.hk)}
\thanks{Lingyu Qiu is with the Department of Mathematics and Applications, University of Naples Federico II, Naples, Italy (e-mail: qiulingyu@gmail.com).}%
\thanks{Yige Li is with the School of Computing and Information Systems, Singapore Management University, Singapore (e-mail: yigeli@smu.edu.sg).}%
\thanks{Qi Yi, and Jitao Sang are with the School of Computer and Information Technology and the Beijing Key Laboratory of Traffic Data Analysis and Mining, Beijing Jiaotong University, Beijing, China (e-mail: 21125273@bjtu.edu.cn, jtsang@bjtu.edu.cn). }%
\thanks{Changsheng Xu is with the National Lab of Pattern Recognition, Institute of Automation, Chinese Academy of Sciences, Beijing, China, and also with the University of Chinese Academy of Sciences, Beijing 100190, China (e-mail: csxu@nlpr.ia.ac.cn).}%

}




\maketitle

\begin{abstract}
The vulnerability of Deep Neural Networks (DNNs) to adversarial attacks poses a significant challenge to their deployment in safety-critical applications. 
While extensive research has addressed various attack scenarios, the no-box attack setting—where adversaries have no prior knowledge, including access to training data of the target model—remains relatively underexplored despite its practical relevance.
This work presents a systematic investigation into leveraging large-scale Vision-Language Models (VLMs), particularly CLIP, as surrogate models for executing no-box attacks.
Our theoretical and empirical analyses reveal a key limitation in the execution of no-box attacks stemming from insufficient discriminative capabilities for direct application of vanilla CLIP as a surrogate model .
To address this limitation, we propose \textbf{MF-CLIP} (Margin-based Fine-tuned CLIP), a novel framework that enhances CLIP's effectiveness as a surrogate model through margin-aware feature space optimization.
Comprehensive evaluations across diverse architectures and datasets demonstrate that MF-CLIP substantially advances the state-of-the-art in no-box attacks, surpassing existing baselines by \textbf{15.23\%} on standard models and achieving a \textbf{9.52\%} improvement on adversarially trained models.
Our code (attached in the supplement) will be made publicly available to facilitate reproducibility and future research in this direction. 
\end{abstract}

\begin{IEEEkeywords}
Adversarial attacks, vision-language model
\end{IEEEkeywords}

\section{Introduction}

\IEEEPARstart{D}{eep} Neural Networks (DNNs) have demonstrated exceptional performance across various applications, demonstrating notable generalization capabilities. Despite these advancements, DNNs remain vulnerable to adversarial attacks~\cite{szegedy2013intriguing}. These attacks involve subtle, often imperceptible perturbations to input images that exploit inherent weaknesses in DNNs' decision boundaries across multiple attack scenarios: white-box, black-box, and no-box. Each scenario introduces unique challenges based on the attacker's level of knowledge about the target model, forming a continuum of decreasing information accessibility.

Among these, the no-box scenario, which lies at the extreme end of the knowledge spectrum, presents a particularly challenging threat model. In this setting, the attacker has zero knowledge about the target model—lacking access to its architecture, parameters, training methodology, and crucially, training data~\cite{li2020practical}. This distinguishes no-box attacks from the more commonly studied black-box attacks, where adversaries typically have at least partial knowledge about training data or can query the model for feedback. The no-box scenario represents a more realistic security concern in production environments where models are completely inaccessible behind APIs or embedded systems, offering a practical reflection of real-world security challenges in deployed machine learning systems.

Despite its practical significance, executing effective no-box attacks remains challenging due to the fundamental issue of transferability—the ability of adversarial examples crafted using one model to successfully deceive another. This transferability becomes especially difficult when operating with minimal knowledge about the target model. The effectiveness of no-box attacks hinges critically on the choice and configuration of surrogate models, which serve as proxies for generating adversarial examples in the absence of direct access to target models. However, this crucial aspect of surrogate model selection has been largely overlooked in existing literature, with most research focusing instead on sophisticated attack methodologies rather than the foundation upon which these attacks are built.

To address this gap, we establish comprehensive criteria for selecting and optimizing surrogate models specifically tailored for no-box adversarial attacks. Our approach reconceptualizes adversarial attacks as an image generation problem, specifically aiming to generate non-robust features that can effectively transfer across model boundaries. Drawing from established principles in image generation~\cite{ramesh2022hierarchical}, we formulate a core principle: the ideal surrogate model should be trained on a diverse and comprehensive dataset that approximates the knowledge spectrum of potential target models, allowing it to generate perturbations that exploit common vulnerabilities across various architectures.

In this context, large-scale Vision-Language Models (VLMs) emerge as promising candidates, with CLIP~\cite{radford2021learning} standing out for its exceptional versatility and rich representational capacity\footnote{For simplicity, ``CLIP" refers to CLIP's image encoder unless otherwise noted throughout this paper.}. CLIP's training on massive and diverse image-text pairs positions it as an ideal surrogate model candidate for transferable attacks. However, our initial experiments reveal a counterintuitive finding: the direct application of vanilla CLIP as a surrogate model fails to yield improved performance, challenging the intuitive expectation (Section~\ref{further}). Through rigorous analysis, we discover that while CLIP possesses remarkable ability to represent diverse visual features across domains, it lacks the necessary discriminative capability essential for effective no-box adversarial attacks—a finding that reveals a fundamental limitation in using foundation models as-is for adversarial purposes.

To further investigate this phenomenon, we conduct a theoretical analysis examining the relationship between discriminative capacity and the inter-class gap, quantifiable by a metric we term the margin. This analysis, complemented by visual examination of CLIP's feature space distribution, confirms that while CLIP excels in representational capacity, it exhibits significant limitations in discriminative power. Specifically, CLIP's feature space reveals compressed inter-class distances within specific domains, creating a fundamental barrier to generating effective domain-specific adversarial perturbations.

\begin{figure}[t]
  \centering
  \includegraphics[width=\linewidth]{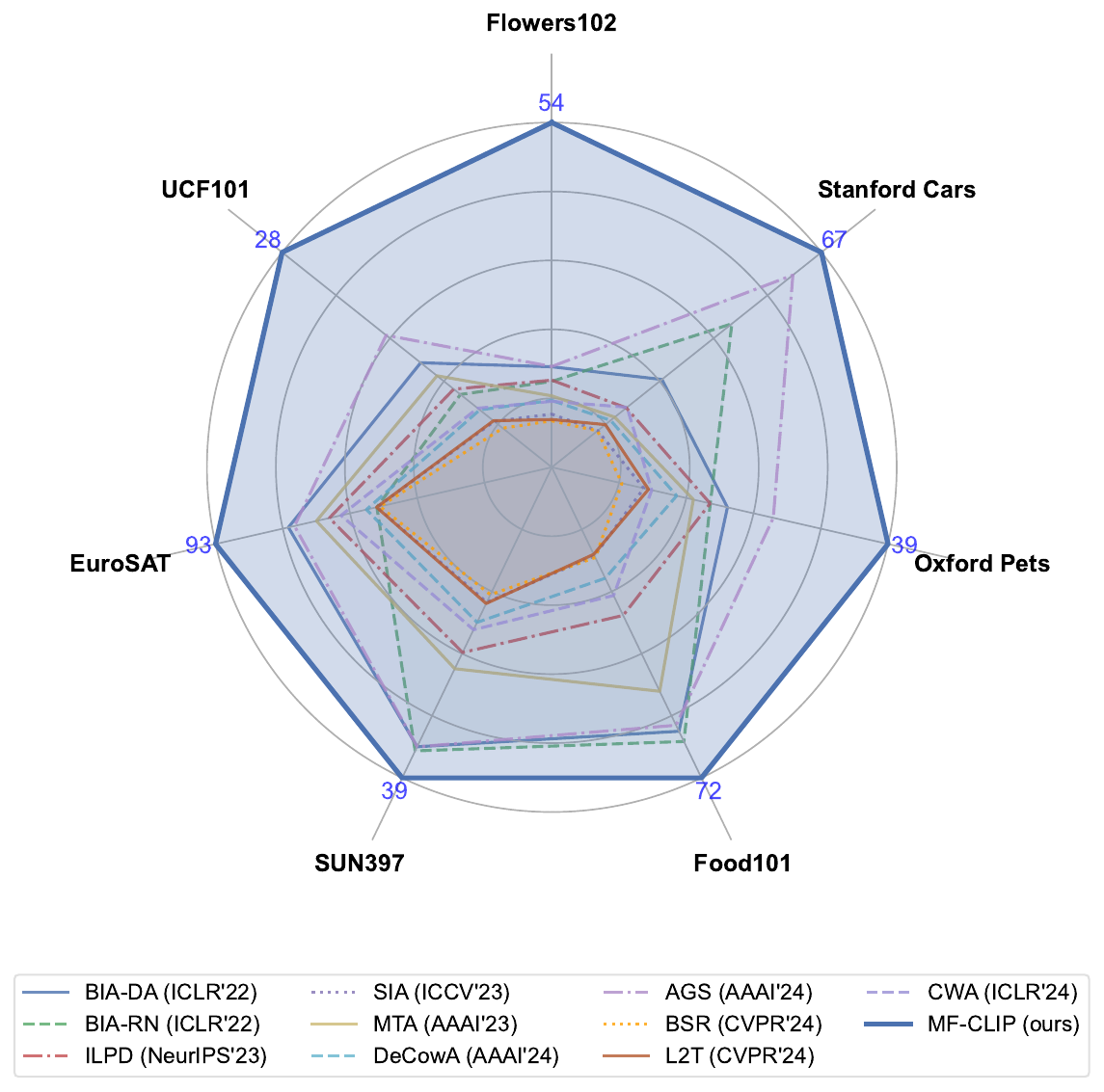}
  \caption{Comparing MF-CLIP's attack success rate (ASR) against state-of-the-art methods across seven datasets. Results represent average performance across three target models (EfficientNet-B0, RegNetX-1.6GF, and ResNet-18). The visualization demonstrates MF-CLIP's consistent and substantial performance advantages across all datasets.}
  \label{fig:radar_plot}
\end{figure}

Building on these insights, we propose a novel fine-tuning methodology named \textbf{Margin-based Fine-tuned CLIP (MF-CLIP)}. This approach is specifically designed to enhance CLIP's discriminative abilities through targeted margin optimization, transforming it into an efficient surrogate model for no-box adversarial attack scenarios. Crucially, the images used in this fine-tuning process—which we refer to as target images—have no overlap with the training sets of the corresponding target models, ensuring strict compliance with the assumptions of the no-box scenario.

Our comprehensive experimental evaluation across diverse datasets and target models demonstrates that MF-CLIP substantially advances the state-of-the-art in no-box adversarial attacks, outperforming the best baseline method by an average of \textbf{15.23\%} on standard models (Figure~\ref{fig:radar_plot}) and achieving a \textbf{9.52\%} improvement on adversarially trained models. These results not only validate our approach but also highlight the critical, yet often overlooked, role of surrogate model optimization in adversarial machine learning.

Our main contributions are summarized as follows:

\begin{itemize}
    \item We provide a systematic investigation into the often-overlooked role of surrogate models in no-box adversarial attacks, establishing theoretical and empirical foundations for leveraging large-scale VLMs with diverse data exposure to improve attack efficacy in challenging scenarios.

    \item We propose MF-CLIP, a margin-based fine-tuned version of CLIP that addresses the fundamental limitation of compressed inter-class distances in foundation models, thereby enhancing discriminative power for domain-specific adversarial attacks while preserving CLIP's rich representational capabilities.

    \item We demonstrate through extensive experimentation the superiority of MF-CLIP over existing attack methods across a range of models, datasets, and defensive strategies, offering novel insights into the utility of surrogate model fine-tuning in adversarial settings and establishing new benchmarks for no-box attacks.
\end{itemize}

\section{Background and Related Work}\label{sec:releated}

\subsection{Vision-Language Models}

Recent advances in Natural Language Processing (NLP) have led to the creation of large-scale models that enjoy a broad spectrum of knowledge~\cite{bommasani2021opportunities}. In computer vision (CV), the focus has shifted toward developing multimodal large-scale models that integrate visual and textual information. While some models, such as Qwen 2.5~\cite{yang2024qwen2} and Llava 1.5~\cite{liu2024improved}, NLP-centric models to support multimodal functionalities, they remain predominantly optimized for text. In contrast, CLIP~\cite{radford2021learning} achieves a more balanced integration of visual and textual modalities, as demonstrated by the widespread standalone use of its image encoder in various CV tasks. This highlights CLIP's effectiveness as a bridge between language and vision.

\subsection{Adversarial Attacks: Evolution and Classification}
 
Adversarial attacks, first identified by Szegedy et al.~\cite{szegedy2013intriguing}, exploit vulnerabilities in neural networks by introducing carefully crafted perturbations to input data that cause misclassification while remaining imperceptible to human observers. These attacks can be broadly categorized based on implementation strategy (query-based versus transfer-based) and knowledge assumptions (white-box, black-box, and no-box).

Query-based attacks rely on iterative feedback from the target model to refine adversarial examples, requiring direct access to model outputs. In contrast, transfer-based attacks—the focus of this paper—generate adversarial examples using a surrogate model and exploit the transferability property to attack target models without requiring query access. This approach is particularly relevant in real-world scenarios where direct interaction with target models may be restricted or monitored.

The evolution of transfer-based attacks can be delineated into three distinct phases, representing a continuum of decreasing knowledge about the target model:

\begin{itemize}
    \item \textbf{White-box Setting}: The adversary has complete access to the target model's architecture, parameters, and training data. This represents the strongest attack scenario, where adversarial examples can be optimized directly against the target model~\cite{goodfellow2014explaining, madry2017towards}.

    \item \textbf{Black-box Setting}: The adversary has no access to the model's architecture or parameters but retains knowledge about or access to the training data distribution. This scenario has received extensive research attention, with numerous techniques developed to enhance transferability across architectures~\cite{dong2018boosting, xie2019improving}.

    \item \textbf{No-box Setting}: Introduced by Li et al.~\cite{li2020practical}, this represents the most challenging scenario where the adversary lacks access to the model's architecture, parameters, and crucially, its training data. This setting models realistic security threats against deployed systems where attackers have minimal information about the target model.
\end{itemize}

The progression through these phases reflects increasing practical relevance but also growing technical difficulty, as each reduction in knowledge significantly constrains the attacker's ability to craft effective adversarial examples.

\subsection{Adversarial Attack Methodologies}

Recent advancements in adversarial attacks have explored various settings, from white-box to no-box scenarios, with each introducing unique challenges and requiring innovative solutions. To address these challenges, several distinct approaches have emerged in the literature:
\begin{itemize}[left=0pt]
  \item \textbf{Gradient Enhancement Methods}: Techniques like MI-FGSM~\cite{dong2018boosting} improve gradients to improve adversarial effectiveness.
  
  \item \textbf{Input Augmentation Strategies}: Methods such as SIA~\cite{wang2023structure}, DeCowA~\cite{lin2024boosting}, L2T~\cite{zhu2024learning}, and BSR~\cite{wang2024boosting} apply input transformations to improve attack generalization across models.
  
  \item \textbf{Intermediate Layer Attacks}: These approaches target intermediate layers in surrogate models, often achieving superior transferability compared to attacks on the final layer. Notable methods include ILA~\cite{huang2019enhancing}, FIA~\cite{wang2021feature}, and ILPD~\cite{li2023improving}.
  
  \item \textbf{Generator-based Approaches}: Generator models, such as BIA~\cite{zhang2022beyond}, are used to produce adversarial examples directly.
  
  \item \textbf{Layer Modification Techniques}: Strategies like AGS~\cite{wang2024ags} modify specific layers within surrogate models to boost attack effectiveness.
\end{itemize}
These categories represent a spectrum of innovative approaches that address the evolving challenges of adversarial attacks in various access settings. 
Our proposed method emphasizes fine-tuning CLIP and employs a generator model to generate adversarial images.

\section{Proposed Method}

In this section, we first establish the theoretical foundations of our approach by outlining the threat model and problem formulation. We then provide a detailed analysis of CLIP's capabilities and limitations in the context of adversarial attacks, culminating in the introduction of our proposed method, MF-CLIP.

\subsection{Preliminaries}
\paragraph{Threat Model}
Our work addresses a practical no-box attack setting that realistically models security threats against deployed machine learning systems. In this setting, the adversary has access to a surrogate model but possesses no information about the target model's architecture, parameters, or—crucially—its training data distribution. This represents a significant departure from conventional black-box settings, where knowledge of or access to the target's training distribution is typically assumed.

The no-box setting creates a substantial knowledge gap between surrogate and target models, making the transferability of adversarial examples particularly challenging. The attacker must indirectly compromise the target using only the surrogate model as a proxy, though they can leverage additional data (explicitly guaranteed not to overlap with the target model's training set) to enhance their attack strategy. This model represents real-world scenarios where attackers might target commercially deployed systems without access to proprietary training data or model details.

\paragraph{Formulation}
We formally define the no-box attack scenario as follows: Given a set of original images \( x \) with corresponding ground truth labels \( y \), we aim to craft adversarial examples \( x' \) using a pre-trained surrogate model \( f \). The objective is to generate \( x' \) that are visually indistinguishable from \( x \) (preserving semantic content) while inducing misclassification in the target model \( g \), such that \( g(x') \neq y \).

We focus on untargeted attacks, where the goal is to cause any misclassification rather than forcing a specific target class. Additionally, we constrain the adversarial perturbations \( \delta = x' - x \) to satisfy \( \| \delta \|_\infty \leq \epsilon \), where \( \epsilon \) represents a pre-defined bound on the perturbation magnitude, ensuring that the adversarial examples remain visually similar to the original images.

For the surrogate model \( f \) with parameters \( \theta_f \), the objective is to maximize its loss function \( J(x', y) \) with respect to the adversarial examples \( x' \):
\begin{equation}
\arg \max_{x'} J(x', y; \theta_f), \quad \text{s.t.} \quad \| x' - x \|_{\infty} \leq \epsilon.
\end{equation}

To efficiently generate these adversarial examples, we employ a generator network \( G \) trained to produce perturbations that maximize the surrogate model's loss. The adversarial examples are then obtained as \( x' = G(x) \), with the generator implicitly enforcing the perturbation constraint during training.

\subsection{Analysis of CLIP's Representational and Discriminative Capabilities}

Before introducing our proposed method, we first analyze CLIP's fundamental properties to understand its potential and limitations as a surrogate model for adversarial attacks. This analysis provides the theoretical foundation for our approach.

\paragraph{Overview of CLIP Architecture and Training Methodology}
CLIP (Contrastive Language-Image Pre-training) comprises two parallel encoders: an image encoder \( f_{\psi} \) and a text encoder \( f_{\phi} \). The image encoder \( f_{\psi} \) maps input images \( x \) to dense embeddings \( f_{\psi}(x) \in \mathbb{R}^d \), while the text encoder \( f_{\phi} \) projects textual descriptions \( y \) to the same embedding space \( f_{\phi}(y) \in \mathbb{R}^d \). 

CLIP employs a similarity function \( h(x,y) = \text{sim}(f_{\psi}(x),f_{\phi}(y)) \) to measure alignment between visual and textual representations. This similarity is defined as the normalized dot product (cosine similarity) of the respective embeddings:
\begin{equation}
h(x,y)=\frac{\langle f_{\psi}(x), f_{\phi}(y) \rangle}{\| f_{\psi}(x) \| \| f_{\phi}(y) \|}.
\end{equation}

During training, CLIP optimizes a contrastive loss function with a temperature parameter \( \tau \) across mini-batches of size \( N \). This loss function encourages matching image-text pairs to have high similarity while minimizing similarity between non-matching pairs:
\begin{equation}
\begin{aligned}
L_S &=\frac{1}{N}\sum_{i=1}^{N}{-\log(\frac{\exp(h(x_i,y_i)/ \tau)}{\sum_{j=1}^{N}\exp(h(x_j,y_i)/ \tau)}})\\
&\quad +\frac{1}{N}\sum_{i=1}^{N}{-\log(\frac{\exp(h(x_i,y_i)/ \tau)}{\sum_{j=1}^{N}\exp(h(x_i,y_j)/ \tau)})}.
\end{aligned}
\label{equation2a}
\end{equation}

This can be reformulated to highlight the relative differences in similarities:
\begin{equation}
\begin{aligned}
L_S &=\frac{1}{N}\sum_{i=1}^{N}{\log(\sum_{j=1}^{N}\exp([h(x_j,y_i)-h(x_i,y_i)]/ \tau))} \\
&\quad +\frac{1}{N}\sum_{i=1}^{N}{\log(\sum_{j=1}^{N}\exp([h(x_i,y_j)-h(x_i,y_i)]/ \tau))}.
\end{aligned}
\label{equation2b}
\end{equation}

\paragraph{Theoretical Analysis of Margin in CLIP's Feature Space}

The inter-class margin—the separation between different classes in feature space—plays a critical role in determining a model's discriminative power and, consequently, its effectiveness in classification tasks. In the context of adversarial attacks, models with smaller margins between classes are generally more vulnerable, as smaller perturbations can cross decision boundaries.

For machine learning models, the margin quantifies the inherent separation between distinct categories within the feature space~\cite{deng2019arcface}. It represents the difference in similarity between positive pairs (samples from the same class) and negative pairs (samples from different classes), typically expressed as \( h(x, x^{+}) - h(x, x^{-}) \), where \( h \) denotes the similarity function.

In CLIP's feature space, for an image \( x \) with true label \( y \) and a different label \( y' \), the margin \( \Delta \) can be mathematically formulated as:
\begin{equation}
\begin{aligned}
    \Delta
    &= h(x,y) - h(x,y') \\
    &= \frac{1}{2} \left( 2 - \|f_{\psi}(x) - f_{\phi}(y)\|^2 \right) \\
    &- \frac{1}{2} \left( 2 - \|f_{\psi}(x) - f_{\phi}(y')\|^2 \right) \\
    &= \frac{1}{2} \left( \|f_{\psi}(x) - f_{\phi}(y')\|^2 - \|f_{\psi}(x) - f_{\phi}(y)\|^2 \right).
\end{aligned} 
\label{marginh}
\end{equation}

This formulation reveals that CLIP's contrastive loss implicitly optimizes the margin between classes. By rewriting Equation \eqref{equation2b} in terms of Euclidean distances in the feature space:
\begin{equation}
\begin{aligned}
    L_S
    =\frac{1}{2N}\sum_{i=1}^{N}{\log(\sum_{j=1}^{N}\exp([\| f_{\psi}(x_i)-f_{\phi}(y_i)\|^{2}} \\
    -{\| f_{\psi}(x_j)-f_{\phi}(y_i)\|^{2}]/ \tau))} \\
    +\frac{1}{2N}\sum_{i=1}^{N}{\log(\sum_{j=1}^{N}\exp([\| f_{\psi}(x_i)-f_{\phi}(y_i)\|^{2}}\\
    - \| f_{\psi}(x_i)-f_{\phi}(y_j)\|^{2}]/ \tau)). \\
\end{aligned}
\label{loss_margin}
\end{equation}

From this reformulation, it becomes evident that CLIP's training objective intrinsically aims to maximize the margin between classes in its feature space. However, our analysis reveals a critical limitation: while CLIP optimizes margins across its entire training distribution of 400 million image-text pairs, the resulting margins for specific downstream domains may be suboptimal due to the generalist nature of its training.

\begin{figure}[b]
  \centering
  \begin{minipage}{0.48\linewidth}
    \includegraphics[width=\linewidth]{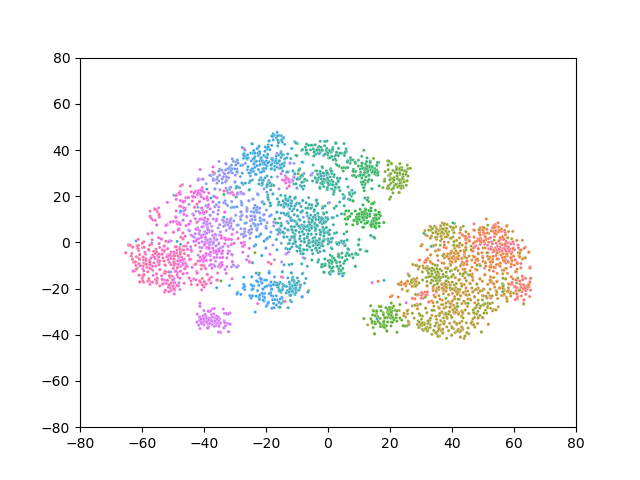} % 替换为你的图片文件路径
    \centerline{(a) Vanilla CLIP}
  \end{minipage}\hfill
  \begin{minipage}{0.48\linewidth}
    \includegraphics[width=\linewidth]{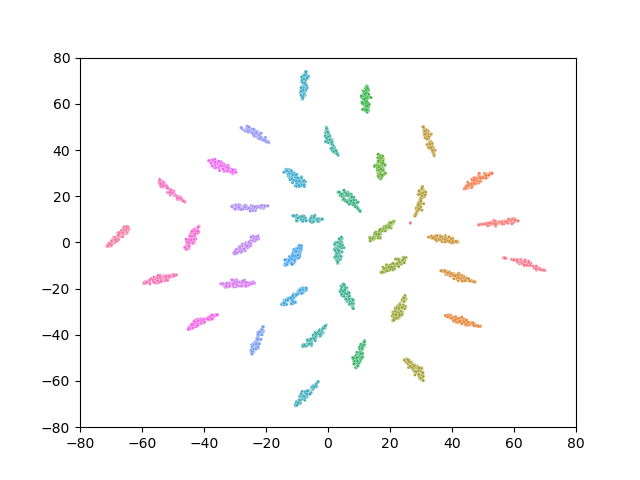}
    \centerline{(b) MF-CLIP}
  \end{minipage}
  \caption{The t-SNE visualization results for the (a) vanilla CLIP and (b) MF-CLIP on the 37-class OxfordPet dataset.}
  \label{fig2}
\end{figure}

\paragraph{Empirical Analysis through Feature Space Visualization}

To complement our theoretical analysis with empirical evidence, we visualize CLIP's feature space using t-SNE~\cite{van2008visualizing}, as illustrated in Figure~\ref{fig2}. Focusing on the 37-class OxfordPet dataset, our visualization reveals two key insights:

1. \textbf{Strong Representational Capacity}: CLIP's embeddings naturally organize into two distinct high-level clusters corresponding to cats and dogs, demonstrating the model's ability to capture broad semantic relationships without fine-tuning. This confirms CLIP's exceptional representational capacity, which allows it to effectively model diverse visual concepts.

2. \textbf{Limited Discriminative Power}: Despite successfully separating high-level categories, CLIP exhibits narrow margins between neighboring fine-grained classes (specific breeds) within each cluster. This compressed inter-class spacing reveals a critical limitation in CLIP's discriminative capabilities for fine-grained classification tasks.

This visualization confirms our theoretical hypothesis: while CLIP excels at representing broad semantic relationships due to its diverse training, it lacks the fine-grained discriminative capabilities necessary for specific downstream tasks. In the context of adversarial attacks, this limitation is particularly problematic, as effective attacks require exploiting precise decision boundaries between classes—boundaries that are poorly defined when inter-class margins are compressed.

The insufficient discriminative power within specific domains represents a fundamental barrier to CLIP's effectiveness as a surrogate model for no-box adversarial attacks. This finding motivates our proposed approach: enhancing CLIP's discriminative capabilities through targeted fine-tuning while preserving its rich representational power.

\subsection{MF-CLIP: Enhancing CLIP's Discriminative Capabilities}

Building on our theoretical and empirical analysis, we propose Margin-based Fine-tuned CLIP (\textbf{MF-CLIP}), a novel approach designed to enhance CLIP's effectiveness as a surrogate model for no-box adversarial attacks by specifically addressing its limited discriminative capabilities.

The key insight driving our approach is that while CLIP possesses exceptional representational capacity due to its training on diverse data, its discriminative power within specific domains—crucial for generating effective adversarial examples—requires enhancement. MF-CLIP achieves this through a targeted fine-tuning process that explicitly optimizes inter-class margins while preserving CLIP's rich representational capabilities.

\begin{figure*}[t]
  \centering
  \includegraphics[width=\linewidth]{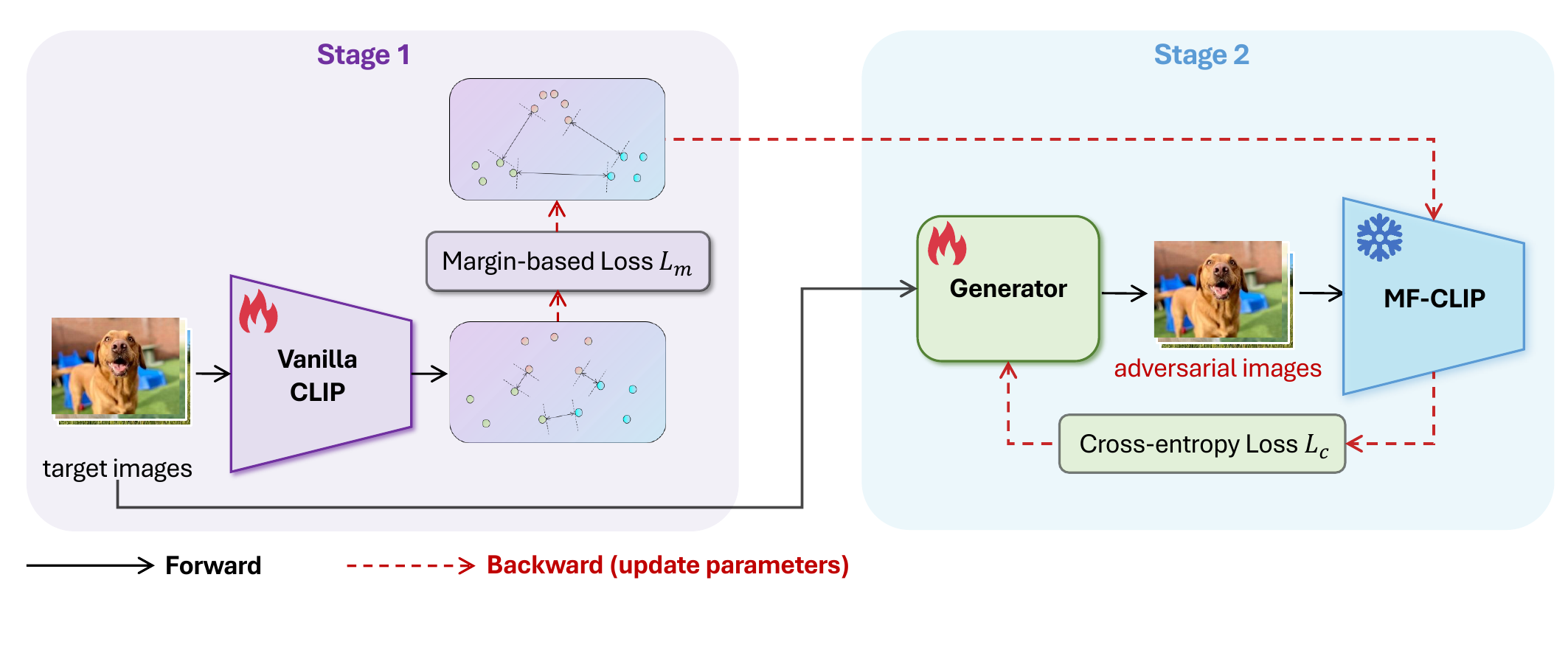}
  \caption{The proposed MF-CLIP framework. Our approach consists of two main stages: (1) Margin-aware fine-tuning, which enhances CLIP's discriminative capabilities by optimizing inter-class margins while preserving its representational power; and (2) Adversarial example generation, where the fine-tuned model serves as a surrogate to generate highly transferable adversarial examples. In the diagram, black solid lines represent the forward process, while red dashed lines indicate the backward process (parameter updates).}
  \label{fig:framework}
\end{figure*}

Figure~\ref{fig:framework} illustrates our proposed two-stage pipeline for MF-CLIP. In the first stage, we enhance CLIP's discriminative capabilities through margin-aware fine-tuning. In the second stage, we leverage the enhanced model as a surrogate to generate transferable adversarial examples.

\paragraph{Stage 1: Margin-Aware Fine-Tuning}

In the first stage, we utilize CLIP as a feature extractor \(f_\theta\) with parameters \(\theta\) to extract \(d\)-dimensional embeddings from input images \(x\). To enhance its discriminative capabilities, we augment CLIP with a supervisory head \(h\), implemented as a fully connected network, and fine-tune the entire model using a margin-based loss function.

Specifically, we employ the Additive Angular Margin Loss~\cite{deng2019arcface}, a state-of-the-art approach for enhancing inter-class separability in embedding spaces:
\begin{equation}
\label{eq1}
L_m=-\log\frac{e^{f_{{\theta}}(x_i) (\cos(\alpha_{y_i}+m))}}{e^{f_{{\theta}}(x_i)(\cos(\alpha_{y_i}+m))}+\sum_{j=1,j\neq y_i}^{n}e^{f_{{\theta}}(x_i)\cos\alpha_{j}}},
\end{equation}

where \(h\) is parameterized by weights \(W\) and a bias \(b\) (set to 0). Here, \(W_j\) denotes the \(j\)-th column of \(W\), \(f_{{\theta}}(x_i)\) represents the embedding of the \(i\)-th image from the \(y_i\)-th class, and \(n\) indicates the total number of classes in the target dataset.

With \(b=0\), the logit can be expressed as \(W^T_j f_{{\theta}}(x_i) = \| W_j \| \| f_{{\theta}}(x_i) \|\cos\alpha_j\), where \(\alpha_j\) represents the angle between weight vector \(W_j\) and feature vector \(f_{{\theta}}(x_i)\). The margin penalty \(m\) explicitly increases the angular separation between classes, enhancing the model's discriminative power.

Crucially, to adhere to the constraints of the no-box setting, we fine-tune using target images \(x\) that are specifically selected for the attack and guaranteed to have no overlap with the training sets of the target models. This ensures that our approach remains applicable in realistic scenarios where attackers have no access to the target model's training data.

\paragraph{Stage 2: Adversarial Example Generation}

In the second stage, we employ the fine-tuned CLIP model \(f_{\hat{\theta}}\) as a surrogate for generating adversarial examples. Rather than directly optimizing perturbations through iterative methods, we train a generator network \(G\) to produce adversarial examples efficiently.

The generator \(G\) is based on a U-Net architecture, incorporating downsampling and upsampling layers with skip connections and ResBlocks. This design allows the generator to learn complex perturbation patterns while maintaining the spatial structure of the input images. The generator is trained to induce misclassification in the fine-tuned surrogate model:
\begin{equation}
\label{eq_c}
L_c = -\mathcal{C} (F(G(x)), y), \quad \text{s.t.} \quad \| G(x) - x \|_{\infty} \leq \epsilon,
\end{equation}

where \(\mathcal{C}\) denotes the cross-entropy loss, and \(F = f_{\hat{\theta}} \circ h\) represents the composite network of the fine-tuned feature extractor and supervisory head. The constraint \(\| G(x) - x \|_{\infty} \leq \epsilon\) ensures that the perturbations remain bounded, preserving the visual similarity between original and adversarial images.

By combining margin-aware fine-tuning with efficient adversarial example generation, MF-CLIP addresses the fundamental limitations of using vanilla CLIP as a surrogate model for no-box attacks. The margin-based fine-tuning enhances CLIP's discriminative capabilities in the target domain, creating more pronounced decision boundaries that can be effectively exploited to generate transferable adversarial examples. Meanwhile, the generator network efficiently learns to produce structured perturbations that maximize the likelihood of successful transfer to unknown target models.

\section{Experiment}\label{sec51}

This section presents comprehensive experimental evaluations of our proposed approach. We first describe our experimental setup, then demonstrate the effectiveness of MF-CLIP through extensive comparisons with state-of-the-art methods on both standard and adversarially trained models, followed by detailed analyses.

\begin{table*}[h!]
    \centering
    \caption{ASR performance comparison of our method against state-of-the-art approaches across various datasets and target models. The higher values indicate better performance. The $\uparrow$ row shows the improvement of our method over the best baseline for each dataset and target model. Average indicates the mean performance of each method across all datasets. The best results are highlighted in \textbf{bold}.}
    \label{table:comprehensive_comparison}
    \setlength{\tabcolsep}{4pt}
    \small
    \begin{tabular}{@{}c l *{8}{C{1.5cm}}@{}}
    \toprule
    \multirow{2}{*}{\rotatebox[origin=c]{90}{\textbf{ }}} & \textbf{Method} & \textbf{Flowers102} & \textbf{Stanford Cars} & \textbf{Oxford Pets} & \textbf{Food101} & \textbf{SUN397} & \textbf{EuroSAT} & \textbf{UCF101} & \textbf{\emph{Average}} \\
    \midrule
    \multirow{12}{*}{\rotatebox[origin=c]{90}{\textbf{EfficientNet-B0}}} 
    & BIA-DA (ICLR 2022) & 15.87 & 19.76 & 23.88 & 60.62 & 37.90 & 70.20 & 19.30 & 35.36 \\
    & BIA-RN (ICLR 2022) & 13.60 & 55.95 & 19.46 & 63.28 & 38.18 & 49.75 & 11.76 & 36.00 \\
    & ILPD (NeurIPS 2023) & 15.87 & 21.94 & 20.93 & 37.04 & 25.49 & 57.46 & 12.19 & 27.27 \\
    & SIA (ICCV 2023) & 10.39 & 14.08 & 13.74 & 22.34 & 19.77 & 43.19 & 8.49 & 18.86 \\
    & MTA (AAAI 2023) & 11.45 & 17.65 & 18.75 & 52.06 & 29.01 & 64.31 & 14.64 & 29.70 \\
    & DeCowA (AAAI 2024) & 12.99 & 18.65 & 16.22 & 28.56 & 22.49 & 47.86 & 9.60 & 22.34 \\
    & AGS (AAAI 2024) & 17.34 & 64.47 & 24.56 & 59.97 & 36.90 & 69.65 & 18.85 & 41.68 \\
    & BSR (CVPR 2024) & 10.35 & 14.08 & 13.98 & 22.14 & 19.59 & 42.06 & 8.19 & 18.63 \\
    & L2T (CVPR 2024) & 10.90 & 16.30 & 14.10 & 21.00 & 21.20 & 44.80 & 8.60 & 19.56 \\
    & CWA (ICLR 2024) & 14.13 & 25.21 & 18.12 & 32.26 & 26.84 & 55.17 & 11.84 & 26.22 \\
    \cmidrule{2-10}
    & \textbf{MF-CLIP (ours)} & \textbf{46.89} & \textbf{66.68} & \textbf{30.50} & \textbf{70.05} & \textbf{40.99} & \textbf{91.83} & \textbf{28.18} & \textbf{53.59} \\
    & $\uparrow$ & +29.55 & +2.21 & +5.94 & +6.77 & +2.81 & +21.63 & +8.88 & +11.91 \\
    \midrule
    \multirow{12}{*}{\rotatebox[origin=c]{90}{\textbf{RegNetX-1.6GF}}} 
    & BIA-DA (ICLR 2022) & 25.46 & 42.18 & 22.95 & 64.93 & 37.24 & 82.79 & 13.53 & 41.30 \\
    & BIA-RN (ICLR 2022) & 20.99 & 51.16 & 22.21 & 69.91 & 37.03 & 63.68 & 12.21 & 39.60 \\
    & ILPD (NeurIPS 2023) & 17.30 & 23.37 & 19.84 & 38.62 & 27.12 & 72.83 & 11.55 & 30.09 \\
    & SIA (ICCV 2023) & 10.80 & 14.53 & 10.60 & 23.94 & 20.84 & 63.69 & 6.66 & 21.58 \\
    & MTA (AAAI 2023) & 15.23 & 18.89 & 17.93 & 57.45 & 25.48 & 72.32 & 10.84 & 31.16 \\
    & DeCowA (AAAI 2024) & 12.42 & 16.83 & 15.21 & 29.09 & 23.20 & 64.10 & 8.67 & 24.22 \\
    & AGS (AAAI 2024) & 15.59 & 64.41 & 26.96 & 59.97 & 36.33 & 74.67 & 14.91 & 41.83 \\
    & BSR (CVPR 2024) & 10.43 & 14.07 & 9.57 & 23.38 & 20.48 & 62.14 & 6.32 & 20.91 \\
    & L2T (CVPR 2024) & 9.20 & 17.00 & 12.50 & 25.00 & 19.60 & 59.60 & 4.80 & 21.10 \\
    & CWA (ICLR 2024) & 14.82 & 24.38 & 14.20 & 34.95 & 27.77 & 71.32 & 9.70 & 28.16 \\
    \cmidrule{2-10}
    & \textbf{MF-CLIP (ours)} & \textbf{61.39} & \textbf{75.60} & \textbf{43.91} & \textbf{74.67} & \textbf{38.89} & \textbf{92.86} & \textbf{28.71} & \textbf{59.43} \\
    & $\uparrow$ & +35.93 & +11.19 & +16.95 & +4.76 & +1.65 & +10.07 & +13.80 & +17.60 \\
    \midrule
    \multirow{12}{*}{\rotatebox[origin=c]{90}{\textbf{ResNet-18}}} 
    & BIA-DA (ICLR 2022) & 6.01 & 19.76 & 13.79 & 57.05 & 31.32 & 65.11 & 8.30 & 28.76 \\
    & BIA-RN (ICLR 2022) & 5.85 & 26.45 & 13.06 & 56.50 & 32.82 & 31.43 & 4.68 & 24.40 \\
    & ILPD (NeurIPS 2023) & 7.88 & 9.97 & 13.90 & 26.83 & 17.94 & 53.95 & 7.01 & 19.64 \\
    & SIA (ICCV 2023) & 3.82 & 5.48 & 7.44 & 15.04 & 10.81 & 39.85 & 2.91 & 12.19 \\
    & MTA (AAAI 2023) & 6.94 & 10.12 & 12.18 & 45.43 & 22.26 & 58.27 & 10.52 & 23.67 \\
    & DeCowA (AAAI 2024) & 6.09 & 8.16 & 11.69 & 19.11 & 13.37 & 41.52 & 4.31 & 14.89 \\
    & AGS (AAAI 2024) & 14.54 & 49.93 & 24.94 & 58.50 & 33.25 & 68.80 & 18.19 & 38.31 \\
    & BSR (CVPR 2024) & 4.10 & 5.75 & 6.62 & 14.90 & 10.46 & 38.17 & 2.88 & 11.84 \\
    & L2T (CVPR 2024) & 4.20 & 8.40 & 8.80 & 14.20 & 13.10 & 40.70 & 4.90 & 13.47 \\
    & CWA (ICLR 2024) & 6.09 & 10.24 & 10.19 & 22.46 & 16.29 & 50.06 & 5.47 & 17.26 \\
    \cmidrule{2-10}
    & \textbf{MF-CLIP (ours)} & \textbf{56.07} & \textbf{55.69} & \textbf{41.67} & \textbf{70.14} & \textbf{38.44} & \textbf{91.77} & \textbf{27.68} & \textbf{54.49} \\
    & $\uparrow$ & +41.53 & +5.76 & +16.73 & +11.64 & +5.62 & +22.97 & +9.49 & +16.18 \\
    \bottomrule
    \end{tabular}
\end{table*}

\begin{table*}[ht!]
    \centering
    \caption{ASR performance comparison of our method against state-of-the-art approaches on the PGD-10 adversarially trained target models. The negative values indicate the adversarial accuracy exceeds the clean accuracy of the adversarially trained models. Average indicates the mean performance of each method across all datasets. The best results are highlighted in \textbf{bold}.}
    \label{table:at_comparison}
    \setlength{\tabcolsep}{4pt}
    \small
    \begin{tabular}{@{}c l *{8}{C{1.5cm}}@{}}
    \toprule
    \multirow{2}{*}{\rotatebox[origin=c]{90}{\textbf{ }}} & \textbf{Method} & \textbf{Flowers102} & \textbf{Stanford Cars} & \textbf{Oxford Pets} & \textbf{Food101} & \textbf{SUN397} & \textbf{EuroSAT} & \textbf{UCF101} & \textbf{\emph{Average}} \\
    \midrule
    \multirow{12}{*}{\rotatebox[origin=c]{90}{\textbf{EfficientNet-B0}}} 
    & BIA-DA (ICLR 2022) & 0.93 & 1.96 & 1.01 & -9.99 & -2.95 & -24.94 & 0.13 & -4.84 \\
    & BIA-RN (ICLR 2022) & 1.02 & 1.75 & 0.74 & -17.53 & -3.41 & -34.79 & -0.11 & -7.48 \\
    & ILPD (NeurIPS 2023) & 0.41 & 1.08 & 0.76 & -12.21 & -3.17 & -31.96 & -0.19 & -6.47 \\
    & SIA (ICCV 2023) & 0.20 & 0.75 & 0.16 & -10.74 & -2.75 & -35.74 & -0.66 & -6.97 \\
    & MTA (AAAI 2023)  & 0.41 & 1.24 & 0.71 & -9.38 & -1.46 & -34.00 & -0.11 & -6.08 \\
    & DeCowA (AAAI 2024) & 0.28 & 0.36 & 0.55 & -10.13 & -3.28 & -36.81 & -0.37 & -7.06 \\
    & AGS (AAAI 2024) & 2.11 & 5.78 & 2.26 & -4.36 & 0.17 & -7.86 & 1.06 & -0.12 \\
    & BSR (CVPR 2024)& 0.08 & 0.46 & 0.41 & -10.62 & -2.87 & -36.36 & -0.42 & -7.05 \\
    & L2T (CVPR 2024) & 1.05 & 0.37 & 0.42 & -11.26 & -1.10 & -32.65 & 0.41 & -6.10 \\
    & CWA (ICLR 2024) & 0.37 & 0.56 & 0.30 & -12.60 & -3.36 & 37.72 & 0.26 & 3.32 \\
    \cmidrule{2-10}
    & \textbf{MF-CLIP (ours)} & \textbf{4.50} & \textbf{12.26} & \textbf{2.32} & \textbf{1.13} & \textbf{2.19} & \textbf{55.85} & \textbf{16.39} & \textbf{13.52} \\
    & $\uparrow$ & +2.39 & +6.48 & +0.06 & +15.49 & +2.02 & +63.71 & +15.33 & +10.20 \\
    \midrule
    \multirow{12}{*}{\rotatebox[origin=c]{90}{\textbf{RegNetX-1.6GF}}} 
    & BIA-DA (ICLR 2022)& 1.02 & 1.54 & 0.87 & -2.74 & 1.23 & 9.31 & 0.00 & 1.60 \\
    & BIA-RN (ICLR 2022)& 0.81 & 1.77 & 1.01 & -2.97 & 2.45 & -14.81 & -0.11 & -1.69 \\
    & ILPD (NeurIPS 2023)& -0.16 & -0.24 & 0.71 & -4.27 & 0.19 & -6.43 & -1.30 & -1.64 \\
    & SIA (ICCV 2023) & -0.16 & -0.24 & 0.71 & -4.27 & 0.19 & -6.43 & -1.30 & -1.64 \\
    & MTA (AAAI 2023) & 0.41 & -0.25 & 0.68 & -3.55 & -0.73 & -1.05 & -0.69 & -0.74 \\
    & DeCowA (AAAI 2024)& -0.37 & -0.22 & 0.57 & -3.90 & -0.70 & -10.68 & -0.95 & -2.32 \\
    & AGS (AAAI 2024) & 1.34 & 7.01 & 1.99 & 1.63 & 6.06 & 10.25 & 4.18 & 4.64 \\
    & BSR (CVPR 2024) & -0.45 & -0.53 & 0.08 & -4.68 & -0.83 & -12.40 & -1.27 & -2.87 \\
    & L2T (CVPR 2024) & 0.52 & -0.74 & 0.16 & -4.48 & 0.07 & -6.60 & -1.00 & -1.58 \\
    & CWA (ICLR 2024) & -0.49 & -0.37 & 0.16 & -4.65 & -3.09 & -5.98& 1.42&  -1.86 \\
    \cmidrule{2-10}
    & \textbf{MF-CLIP (ours)} & \textbf{9.05} & \textbf{12.36} & \textbf{13.16} & \textbf{8.24} & \textbf{13.32} & \textbf{28.12} & \textbf{29.68} & \textbf{14.99} \\
    & $\uparrow$ & +7.71 & +5.35 & +11.17 & +6.61 & +7.26 & +17.87 & +25.50 & +10.35 \\
    \midrule
    \multirow{12}{*}{\rotatebox[origin=c]{90}{\textbf{ResNet-18}}} 
    & BIA-DA (ICLR 2022)& 0.20 & -0.42 & 0.74 & 4.93 & -2.74 & 16.28 & 9.31 & 4.04 \\
    & BIA-RN (ICLR 2022)& 0.20 & -0.67 & 0.68 & 1.58 & -2.97 & -9.68 & -3.09 & -1.99 \\
    & ILPD (NeurIPS 2023) & 0.12 & -1.53 & 0.76 & 0.45 & -1.71 & -2.19 & -2.14 & -0.89 \\
    & SIA (ICCV 2023) & -0.53 & -1.93 & 0.00 & -1.11 & -1.56 & -6.10 & -2.78 & -2.00 \\
    & MTA (AAAI 2023)  & -0.41 & -1.89 & 0.16 & -0.18 & -2.80 & 3.70 & -0.66 & -0.30 \\
    & DeCowA (AAAI 2024)& -0.57 & -1.85 & 0.57 & -0.38 & -1.87 & -6.47 & -2.62 & -1.88 \\
    & AGS (AAAI 2024)& 0.53 & 0.91 & 1.88 & 7.45 & 0.90 & 14.81 & 3.15 & 4.23 \\
    & BSR (CVPR 2024) & -0.49 & -1.64 & 0.11 & -1.29 & -1.66 & -6.70 & -3.01 & -2.10 \\
    & L2T (CVPR 2024) & -0.36 & -2.13 & 0.62 & -1.38 & -0.37 & -4.93 & -2.89 & -1.43 \\
    & CWA (ICLR 2024)& -0.20 & -1.68 & 0.35 & -0.40 & -1.69 & -6.84 & 3.09 & -1.05 \\
    \cmidrule{2-10}
    & \textbf{MF-CLIP (ours)} & \textbf{5.30} & \textbf{1.46} & \textbf{5.89} & \textbf{14.15} & \textbf{4.70} & \textbf{38.7} & \textbf{15.01} & \textbf{12.26} \\
    & $\uparrow$ & +4.77 & +0.55 & +4.01 & +6.70 & +3.80 & +23.89 & +11.86 & +8.03 \\
    \bottomrule
    \end{tabular}
\end{table*}

\subsection{Experiments Settings}

\paragraph{Datasets and Models}
We conduct experiments on seven diverse, high-resolution vision datasets: OxfordPets~\cite{parkhi2012cats}, Flowers102~\cite{nilsback2008automated}, StanfordCars~\cite{krause20133d}, Food101~\cite{bossard2014food}, SUN397~\cite{xiao2010sun}, EuroSAT~\cite{helber2019eurosat}, and UCF101~\cite{soomro2012ucf101}. For target models, we employ ResNet-18~\cite{he2016deep}, EfficientNet-B1~\cite{tan2019efficientnet}, and RegNetX-1.6GF~\cite{radosavovic2020designing}.

\paragraph{Implementation Details}
We fine-tune CLIP using SGD optimizer with a batch size of 128 and an initial learning rate of 0.01, following a cosine annealing schedule over 300 epochs. The generator is trained using AdamW optimizer with a batch size of 64 and an initial learning rate of 0.0001 for 90 epochs, also with cosine annealing. We set the margin parameter $m=0.15$ and constrain the $\ell_\infty$-norm perturbation to $\epsilon=16/255$. Test set images are used as target images unless otherwise specified.

\paragraph{Baselines and Evaluation Metric}
We compare against state-of-the-art methods including AGS~\cite{wang2024ags}, BSR~\cite{wang2024boosting}, CWA~\cite{chenrethinking}, ILPD~\cite{li2023improving}, L2T~\cite{zhu2024learning}, MTA~\cite{qin2023training}, and SIA~\cite{wang2023structure}, implemented using TransferAttack\footnote{\url{https://github.com/Trustworthy-AI-Group/TransferAttack}}. We also include BIA~\cite{zhang2022beyond} using official weights. All methods use ResNet-50 as the surrogate backbone, except CWA (ResNet-50/34 ensemble) and MTA (ResNet-MTA-18). Performance is measured using Attack Success Rate (ASR), defined as the difference between clean and adversarial accuracy. Note that some studies define ASR as 1 minus the adversarial accuracy, which can lead to higher reported ASR values compared to our definition. We choose our definition to provide a more direct measure of the attack's impact by explicitly showing the accuracy degradation.

% \paragraph{Baselines.}
% To demonstrate the effectiveness of our method, we selected a comprehensive set of baseline methods for comparison. Utilizing a PyTorch tool, TransferAttack\footnote{\url{https://github.com/Trustworthy-AI-Group/TransferAttack}}, we deployed AGS~\cite{wang2024ags}, BSR~\cite{wang2024boosting}, CWA~\cite{chenrethinking}, ILPD~\cite{li2023improving}, L2T~\cite{zhu2024learning}, MTA~\cite{qin2023training}, and SIA~\cite{wang2023structure} as baseline methods with default settings. Additionally, we used the officially released weights to deploy BIA~\cite{zhang2022beyond}. 
% Both our method and the baseline methods use ResNet-50 as the surrogate backbone, except for CWA, which uses an ensemble of ResNet-50 and ResNet-34, and MTA, which uses ResNet-MTA-18.

% \paragraph{Metric.}
% The evaluation metric used in this paper is Attack Success Rate (ASR). It is defined as the difference between the accuracy of the target model on clean images and the accuracy on adversarial images. In some other studies, ASR is expressed as 1 minus the accuracy of the target model on adversarial images. This difference in definitions can result in the ASR values presented in this paper being lower than those in other studies.

\subsection{Evaluations on Normal Models}

Table~\ref{table:comprehensive_comparison} presents comprehensive results across all datasets and target models. The row denoted by $\uparrow$ represents the improvement of our method over the best-performing baseline for each scenario. Notably, all values in this row are positive, indicating that MF-CLIP consistently outperforms existing methods across all test conditions. Quantitatively, MF-CLIP achieves average improvements of \textbf{11.91\%}, \textbf{17.60\%}, and \textbf{16.18\%} over the best baseline for EfficientNet-B0, RegNetX-1.6GF, and ResNet-18, respectively. These substantial margins demonstrate the effectiveness of our approach in addressing the challenges of no-box attacks.

\subsection{Evaluations on Adversarially Trained Models}

We further evaluate all methods against models adversarially trained with PGD-10~\cite{madry2017towards} using an $\epsilon$-radius. As shown in Table~\ref{table:at_comparison}, certain baseline methods yield negative ASR values, indicating that the adversarial accuracy surpasses the clean accuracy of the robust models. This outcome aligns with the known trade-off in adversarial training, where increased robustness often leads to a reduction in clean accuracy~\cite{tsiprasrobustness}, highlighting the relative limitations of these attack methods. Notably, even with the enhanced robustness of these target models, MF-CLIP achieves consistent positive gains across all scenarios, outperforming the strongest baseline methods by margins of \textbf{10.20\%}, \textbf{10.35\%}, and \textbf{8.03\%} on EfficientNet-B0, RegNetX-1.6GF, and ResNet-18, respectively. These consistent gains demonstrate the robustness of our approach even against adversarially hardened targets.

% In this subsection, we evaluate the performance of our proposed method and baseline methods on adversarially trained target models. We adversarially trained the target models using PGD-10~\cite{madry2017towards} with $\epsilon$-radius. 
% Similarly to Table~\ref{table:comprehensive_comparison}, we present the comparison results with the baseline methods in Table~\ref{table:at_comparison}.
% Note that the negative values in the table indicate that the clean accuracy of the adversarially trained target models is lower than the accuracy of the adversarial examples generated by the corresponding attack methods. 
% This anomaly arises because adversarial training can lead to overfitting on adversarial examples, thereby increasing adversarial accuracy while decreasing clean accuracy~\cite{tsiprasrobustness}. 
% This also suggests that the corresponding attack methods are relatively weak.
% Although adversarial training can effectively enhance the adversarial robustness of target models, our proposed method still achieves positive improvements across all models and datasets, outperforming all existing methods.

\subsection{Comparison with Fine-tuned Baseline Methods}

\begin{table*}[!t]
    \centering
    \caption{Comparison between MF-CLIP and baseline methods where the surrogate models are replaced with fine-tuned CLIP (when the baseline method supports such modification). Despite these enhanced baselines benefiting from better surrogate models, our method still consistently achieves superior performance across all datasets and target models.}
    \label{table:ft_clip_baselines}
    \begin{adjustbox}{width=\linewidth}
    \begin{tabular}{@{}c l *{7}{C{1.5cm}}@{}}
        \toprule
        \multirow{2}{*}{\rotatebox[origin=c]{90}{\textbf{ }}} & \textbf{Method} & \textbf{Flowers102} & \textbf{Cars} & \textbf{Pets} & \textbf{Food101} & \textbf{SUN397} & \textbf{EuroSAT} & \textbf{UCF101} \\
        \midrule
        \multirow{5}{*}{\rotatebox[origin=c]{90}{\textbf{EfficientNet}}}
        & SIA (ICCV 2023) & 10.08 & 15.41 & 15.62 & 42.87 & 25.08 & 52.19 & 9.81  \\
        & DeCowA (AAAI 2024)  & 13.80 & 18.13 & 17.25 & 47.88 & 25.48 & 62.32 & 11.87  \\
        & BSR (CVPR 2024)  & 10.11 & 15.88 & 15.34 & 40.74 & 24.20 & 56.11 & 9.81  \\
        & L2T (CVPR 2024)  & 11.75 & 14.81 & 16.03 & 46.66 & 26.30 & 56.45 & 9.01  \\
        \cmidrule{2-9}
        & \textbf{MF-CLIP (ours)} & \textbf{46.89} & \textbf{66.68} & \textbf{30.50} & \textbf{70.05} & \textbf{40.99} & \textbf{91.83} & \textbf{28.18} \\
        \midrule
        \multirow{5}{*}{\rotatebox[origin=c]{90}{\textbf{RegNetX}}}
        & SIA (ICCV 2023) & 8.69 & 15.67 & 15.45 & 42.62 & 24.62 & 63.79 & 7.75  \\
        & DeCowA (AAAI 2024) & 13.28 & 17.35 & 19.73 & 49.28 & 26.76 & 63.30 & 11.71  \\
        & BSR (CVPR 2024) & 8.73 & 15.53 & 15.24 & 40.59 & 24.05 & 64.64 & 7.80  \\
        & L2T (CVPR 2024) & 11.05 & 16.93 & 14.45 & 44.85 & 26.37 & 65.52 & 7.29  \\
        \cmidrule{2-9}
        & \textbf{MF-CLIP (ours)} & \textbf{61.39} & \textbf{75.60} & \textbf{43.91} & \textbf{74.67} & \textbf{38.89} & \textbf{92.86} & \textbf{28.71} \\
        \midrule
        \multirow{5}{*}{\rotatebox[origin=c]{90}{\textbf{ResNet-18}}}
        & SIA (ICCV 2023) & 5.03 & 7.65 & 11.28 & 26.80 & 15.23 & 56.62 & 3.57 \\
        & DeCowA (AAAI 2024) & 8.36 & 10.88 & 14.69 & 34.89 & 17.28 & 63.91 & 7.30 \\
        & BSR (CVPR 2024) & 4.95 & 8.31 & 11.94 & 24.89 & 14.53 & 58.74 & 4.28 \\
        & L2T (CVPR 2024) & 5.72 & 8.10 & 13.45 & 30.42 & 16.79 & 60.30 & 3.98  \\
        \cmidrule{2-9}
        & \textbf{MF-CLIP (ours)} & \textbf{56.07} & \textbf{55.69} & \textbf{41.67} & \textbf{70.14} & \textbf{38.44} & \textbf{91.77} & \textbf{27.68} \\
        \bottomrule
    \end{tabular}
    \end{adjustbox}
\end{table*}

To further validate the effectiveness of our approach beyond the surrogate model selection, we conducted an additional experiment where we replaced the surrogate models in baseline methods with fine-tuned CLIP (for those baseline methods that support modifying the surrogate model). This experiment aims to isolate the impact of our margin-based fine-tuning approach from the general benefits of using CLIP as a surrogate model.

As shown in Table~\ref{table:ft_clip_baselines}, even when baseline methods benefit from using fine-tuned CLIP as their surrogate model, MF-CLIP still consistently outperforms all competitors by substantial margins across all datasets and target models. Specifically, on the EfficientNet target model, MF-CLIP achieves improvements of 33.09\% on Flowers102 compared to the best-performing enhanced baseline. Similar patterns are observed for RegNetX and ResNet-18 target models.

These results conclusively demonstrate that the superior performance of MF-CLIP stems not merely from the use of CLIP, but specifically from our margin-based fine-tuning approach that effectively enhances CLIP's discriminative capabilities for adversarial transferability. This finding supports our theoretical analysis that the margin in feature space plays a critical role in determining a model's effectiveness as a surrogate for adversarial attacks.

\subsection{Extended Evaluation on Vision Transformers and ImageNet}

\begin{table*}[!t]
    \centering
    \caption{No-box attack performance on ImageNet validation set across different architectures including Vision Transformers. All methods use ResNet-50 backbone as surrogate model. Results show that MF-CLIP consistently outperforms baselines by large margins, particularly against Vision Transformers.}
    \label{table:imagenet_vit}
    \begin{adjustbox}{width=0.7\linewidth}
    \begin{tabular}{c|cccc}
        \toprule
        \textbf{Method} & \textbf{EfficientNet-B0} & \textbf{RegNetX-1.6GF} & \textbf{ResNet-18} & \textbf{ViT-B/16} \\
        \hline
        SIA (ICCV 2023)  & 22.59  & 25.33  & 24.41  & 6.85  \\
        DeCowA (AAAI 2024)  & 26.22  & 27.86  & 26.90  & 9.80  \\
        BSR (CVPR 2024)  & 24.01  & 27.75  & 25.67  & 6.86  \\
        L2T (CVPR 2024)  & 21.60  & 23.10  & 22.60  & 6.10  \\
        \hline
        \textbf{MF-CLIP (ours)} & \textbf{59.96}  & \textbf{59.40}  & \textbf{57.07}  & \textbf{24.92}  \\
        \bottomrule
    \end{tabular}
    \end{adjustbox}
\end{table*}

To demonstrate the effectiveness in larger datasets and newer architectures like Vision Transformers, we conducted additional experiments on the ImageNet-1K validation set. As shown in Table~\ref{table:imagenet_vit}, we evaluated MF-CLIP against baseline methods on four target architectures: EfficientNet-B0, RegNetX-1.6GF, ResNet-18, and importantly, ViT-B/16. 

The results demonstrate that MF-CLIP substantially outperforms all baseline methods across all architectures. On ImageNet, our method achieves attack success rates of 59.96\%, 59.40\%, and 57.07\% against EfficientNet, RegNetX, and ResNet-18 respectively. This confirms that our approach maintains its effectiveness on large-scale, diverse datasets.

More significantly, MF-CLIP shows impressive transferability to Vision Transformers, achieving a 24.92\% attack success rate against ViT-B/16, more than doubling the best baseline's performance (9.80\%). This result is particularly important because Vision Transformers employ a fundamentally different architecture from convolutional neural networks, with distinct inductive biases and feature extraction mechanisms. The superior performance against ViT-B/16 suggests that MF-CLIP's enhanced discriminative capabilities enable the generation of adversarial examples that exploit vulnerabilities common across diverse architectural paradigms.

These findings conclusively demonstrate that MF-CLIP's effectiveness extends beyond the limited datasets and architectures evaluated in prior work, establishing its utility across both large-scale natural image datasets and modern neural network architectures.

\subsection{Further Analysis}\label{further}

\paragraph{CLIP vs. ImageNet Models}

\begin{figure*}[!ht]
  \centering
  \includegraphics[width=\linewidth]{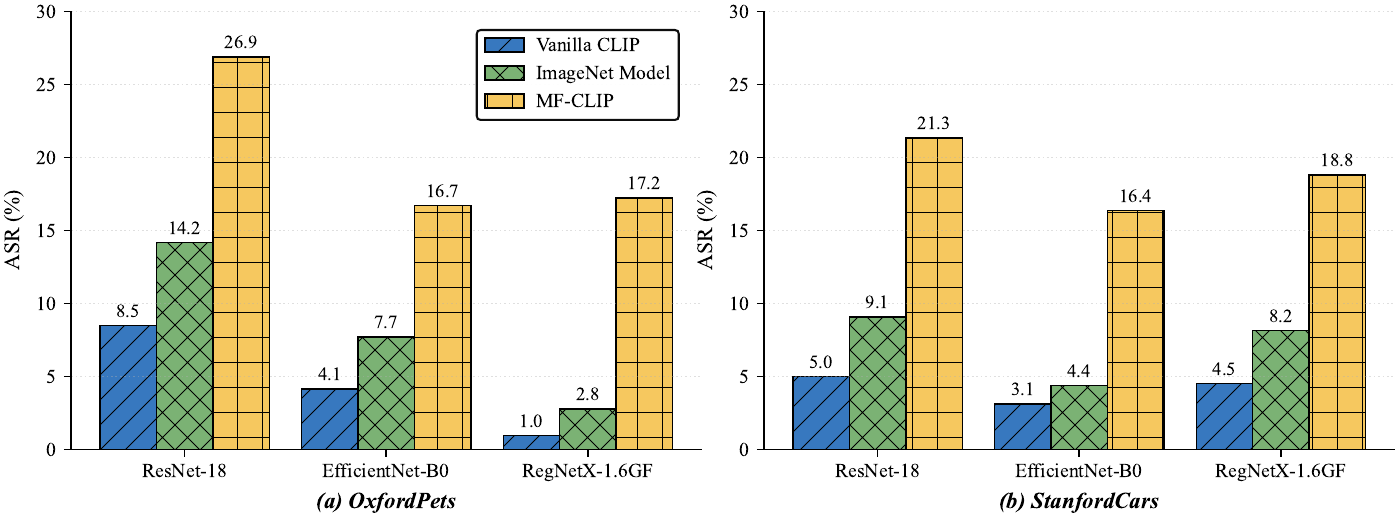} 
  \caption{Analysis for comparing different surrogate models with ResNet-50 backbone. The results clearly demonstrate that unrefined CLIP even performs worse than ImageNet models, while MF-CLIP significantly outperforms both.}
  \label{fig:ablation_clip}
\end{figure*}

% \begin{figure}[!ht]
%   \centering
%   \begin{minipage}{0.95\linewidth}
%     \includegraphics[width=\linewidth]{figs/ig3-car.pdf} % 替换为你的图片文件路径
%     \centerline{(a) StanfordCar}
%   \end{minipage}\hfill
%   \begin{minipage}{0.95\linewidth}
%     \includegraphics[width=\linewidth]{figs/ig3-pets.pdf}
%     \centerline{(b) OxfordPets}
%   \end{minipage}
%   \caption{Comparison of ASR between ImageNet models and CLIP as surrogate models (ResNet-50/ViT-B16) on different target architectures (ResNet-18, EfficientNet-B0, and RegNetX-1.6GF) using StanfordCar and OxfordPets using FGSM attack.}
%   \label{fig:ana}
% \end{figure}

We conduct comprehensive analytical experiments to evaluate vanilla CLIP's suitability as a surrogate model in the no-box setting. Figure~\ref{fig:ablation_clip} presents a critical analysis about comparing different surrogate models with ResNet-50 backbone across multiple target datasets and architectures. The results reveal a surprising and counterintuitive phenomenon: despite CLIP's extensive pre-training on diverse data, vanilla CLIP not only fails to outperform ImageNet models when used as a surrogate, but actually performs worse in generating transferable adversarial examples.

These results provide strong empirical evidence supporting our theoretical analysis regarding CLIP's limited discriminative capabilities within specific domains. They highlight that while CLIP possesses superior representational capacity, this alone is insufficient for effective adversarial transferability. Most importantly, the dramatic performance improvement achieved by MF-CLIP over both vanilla CLIP and ImageNet models (as shown in Figure~\ref{fig:ablation_clip}) demonstrates the effectiveness of our margin-based fine-tuning approach in addressing this limitation, validating our method's core contribution.

\begin{figure}[!ht]
  \centering
  \begin{minipage}{0.95\linewidth}
    \includegraphics[width=\linewidth]{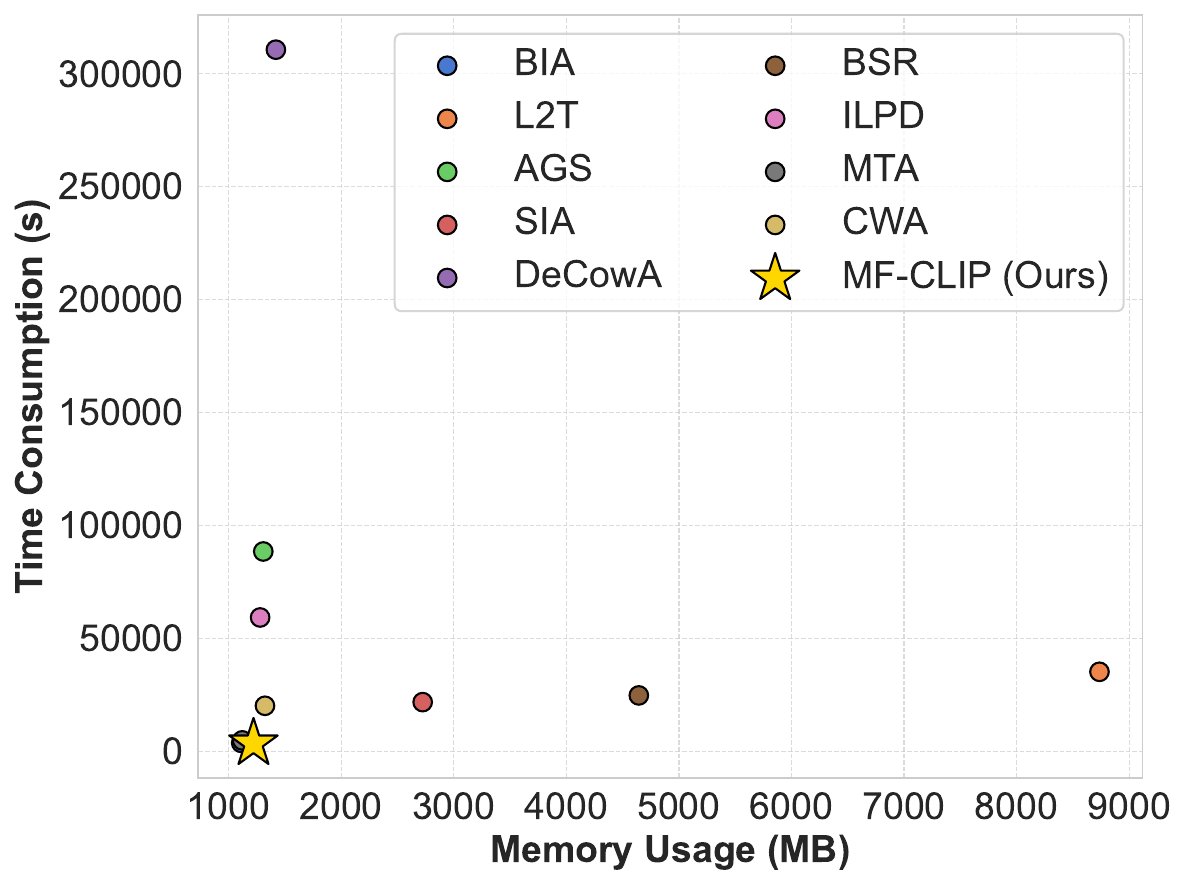} % 替换为你的图片文件路径
    \centerline{(a) Batch size = 1}
  \end{minipage}\hfill
  \begin{minipage}{0.95\linewidth}
    \includegraphics[width=\linewidth]{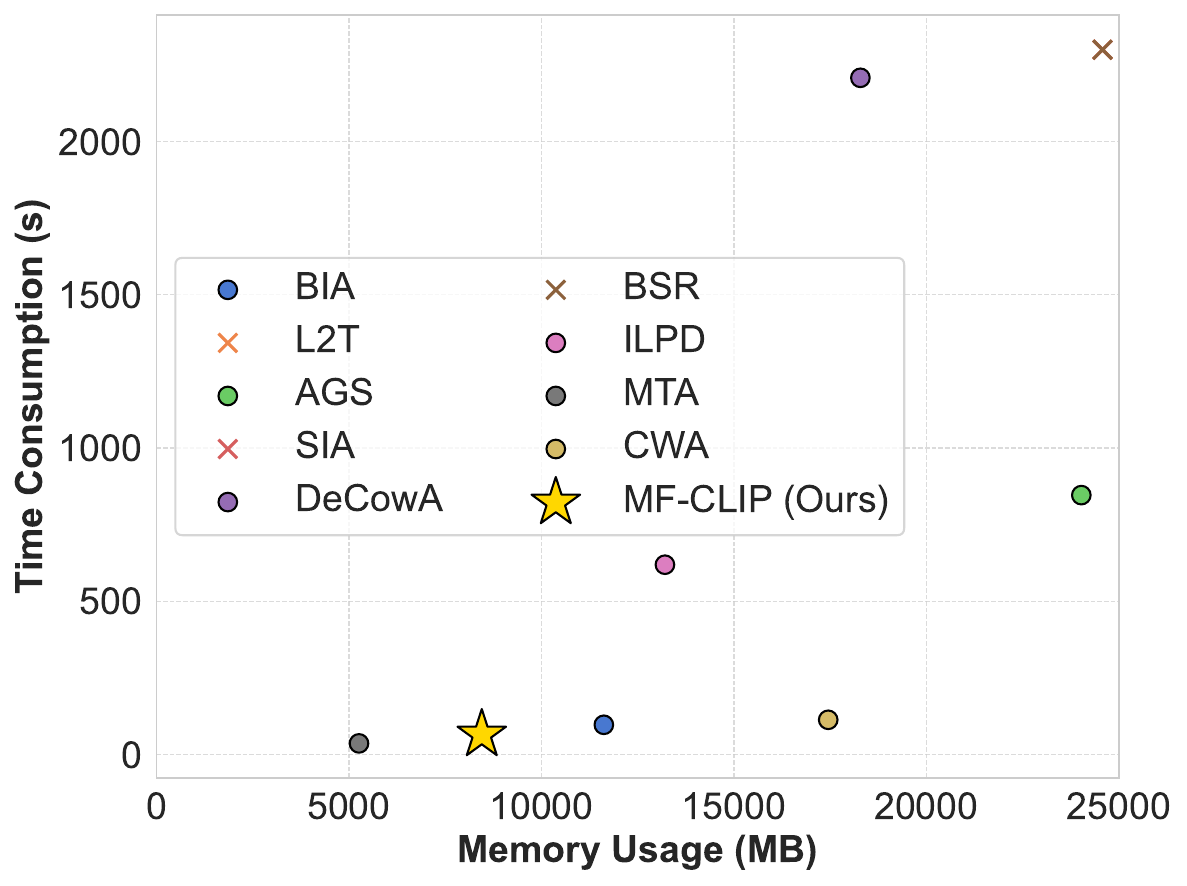}
    \centerline{(b) Batch size = 128}
  \end{minipage}
  \caption{Computational efficiency comparison on Flowers102 dataset. Time consumption vs. memory usage at different batch sizes. The \texttimes\ symbol indicates out of memory failures.}
  \label{fig:time}
\end{figure}

\paragraph{Computational Efficiency}

As shown in Figure~\ref{fig:time}, we evaluate MF-CLIP's efficiency in terms of memory and time costs under two batch size settings. At batch size 1, MF-CLIP demonstrates minimal memory requirements (1,221 MB) and low time cost (3,449 s), significantly outperforming methods like DeCowA and L2T that consume up to 8,736 MB memory and 310,588 seconds. At batch size 128, MF-CLIP maintains efficient scaling with moderate memory usage (8,449 MB) and computation time (about 65 seconds), surpassing most baselines. Notably, in single-inference scenarios (batch size=1), MF-CLIP achieves near-optimal performance in both time and memory efficiency. When operating at batch size 128, while MF-CLIP's memory consumption is slightly higher than BIA, its superior attack performance more than compensates for this marginal overhead. This favorable efficiency profile stems from MF-CLIP's streamlined architecture, making it particularly suitable for scalable, resource-efficient applications.

\paragraph{Ablation Study}

\begin{figure}[!ht]
  \centering
  \includegraphics[width=0.7\linewidth]{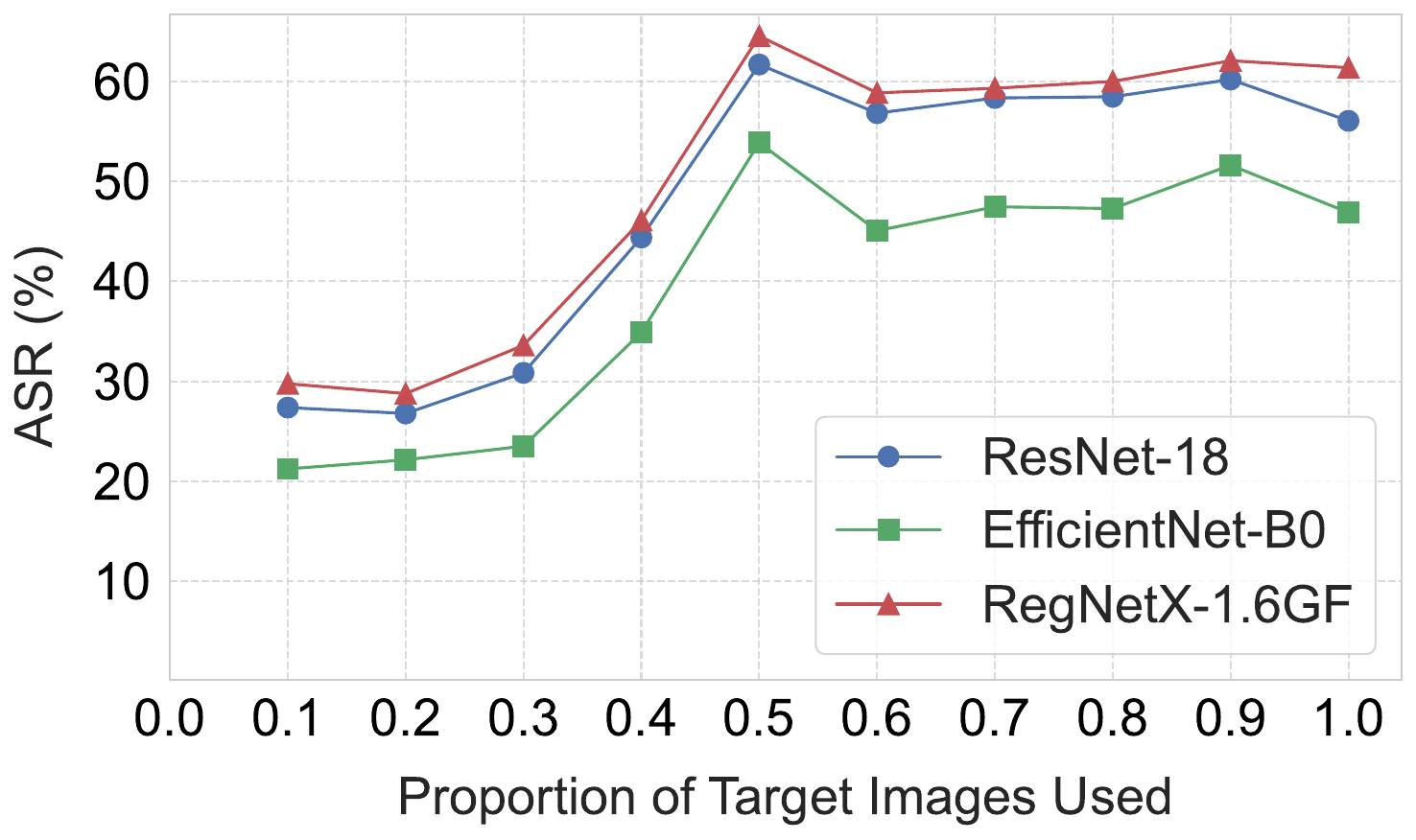}
  \caption{ASR vs. proportion of target images used for fine-tuning on Flowers102 dataset.}\label{fig:number}
\end{figure}

To investigate MF-CLIP's sensitivity to the number of target images available for fine-tuning, we conduct experiments on the Flowers102 dataset using varying proportions (10\% to 100\%) of the test set. As shown in Figure~\ref{fig:number}, while performance generally improves with more target images, MF-CLIP remains remarkably effective even with limited data. Using just 10\% of the target images (199 images across 102 classes, less than 2 images per class), we achieve ASR values of 27.37\%, 21.23\%, and 29.76\% on ResNet-18, EfficientNet-B0, and RegNetX-1.6GF respectively—still outperforming existing methods. These results demonstrate MF-CLIP's robust performance even under significant data constraints.

\section{Conclusion and Discussion}

We presented MF-CLIP, a margin-based fine-tuning approach that enhances CLIP's effectiveness as a surrogate model for no-box adversarial attacks. Our analysis revealed CLIP's limitations in discriminative capabilities, which our method directly addresses. Experiments across diverse datasets, target models, and real-world settings demonstrate that MF-CLIP consistently outperforms state-of-the-art techniques, particularly at lower perturbation budgets.

Our findings reveal that improving a surrogate model's discriminative power is more effective than focusing solely on attack algorithms. MF-CLIP achieves remarkable cross-dataset generalization, maintaining effectiveness against adversarially trained models, Vision Transformers, and commercial API systems. These results provide insights into the relationship between representational capacity, discriminative power, and transferability in neural networks.

Future work may explore integrating MF-CLIP with complementary attack strategies, investigating defensive countermeasures, extending to other modalities, and analyzing the role of margin in adversarial transferability. Beyond adversarial attacks, our findings on improving discriminative power while maintaining representational capacity may have broader implications for transfer learning and foundation model adaptation.

\bibliographystyle{IEEEtran}
\bibliography{main}

\end{document}